%% file: neurips_2025.tex
\documentclass{article}

\usepackage[preprint]{neurips_2025}

\usepackage[utf8]{inputenc} 
\usepackage[T1]{fontenc}    
\usepackage{url}            
\usepackage{booktabs}       
\usepackage{amsfonts}       
\usepackage{nicefrac}       
\usepackage{microtype}      
\usepackage{xcolor}         
\usepackage{graphicx}
\usepackage{multirow}
\usepackage{multicol}
\usepackage{colortbl}
\usepackage{tcolorbox}
\usepackage{subfigure}
\usepackage[subfigure]{tocloft}
\usepackage{adjustbox}

\usepackage[toc,page,header]{appendix}
\usepackage{tocloft}          
\usepackage{minitoc}
\definecolor{deeporange}{rgb}{0.8, 0.3, 0.0}
\usepackage{tikz}
\usepackage{enumitem}
\usepackage{amssymb}
\usepackage{amsmath}
\usepackage{natbib}
\setcitestyle{numbers,square}
\newtheorem{assumption}{Assumption}[section]
\newcommand{\circled}[2][orange]{%
  \tikz[baseline=(char.base)]{
    \node[shape=circle, draw, line width=0.3pt, inner sep=0.5pt, minimum size=6pt, color=#1] (char) {\scriptsize\textcolor{#1}{#2}};
  }%
}

\definecolor{light-gray}{gray}{0.6}
\definecolor{front-color}{HTML}{F5FFFA}
\definecolor{Gray}{gray}{0.93}
\usepackage{xspace}
\newcommand{\eg}{\textit{e.g.}\xspace}

\usepackage{hyperref}
\title{Reinforcing Video Reasoning with Focused Thinking}
\author{
    Jisheng Dang$^{1,2,5}$$^{\dagger}$,\; 
    Jingze Wu$^1$$^{\dagger}$,\;
    Teng Wang$^3$$^*$,\;
    Xuanhui Lin$^2$,\;
    Nannan Zhu$^1$,\; \\
    \textbf{Hongbo Chen}$^1$\textbf{,}\;
    \textbf{Wei-Shi Zheng}$^1$\textbf{,}\;
    \textbf{Meng Wang}$^4$\textbf{,}\;
    \textbf{Tat-Seng Chua}$^5$\\
    $^1$ Sun Yat-sen University
    $^2$ Lanzhou University
    $^3$ University of Hong Kong\\
    $^4$ Hefei University of Technology 
    $^5$ National University of Singapore\\[2mm]
}

\begin{document}
\doparttoc
\faketableofcontents

\maketitle
\let\thefootnote\relax
\footnotetext{$^\dagger$ Equal contribution. $^*$ Corresponding author.}
\input{0-abstract}
\input{1_intro}
\input{2-related-work}
\input{4-method}
\input{5-experiment}
\input{6-conclusion}
\bibliography{Ref}
\bibliographystyle{IEEEtran}
\clearpage
\appendix
\newpage
\addcontentsline{toc}{section}{Appendix}
\part{Appendix}
\parttoc
\input{7-supplementary}


\end{document}

%% file: 0-abstract.tex
\begin{abstract}
Recent advancements in reinforcement learning, particularly through Group Relative Policy Optimization (GRPO), have significantly improved multimodal large language models for complex reasoning tasks. However, two critical limitations persist: 1) they often produce unfocused, verbose reasoning chains that obscure salient spatiotemporal cues and 2) binary rewarding fails to account for partially correct answers, resulting in high reward variance and inefficient learning. In this paper, we propose TW-GRPO, a novel framework that enhances visual reasoning with focused thinking and dense reward granularity. Specifically, we employs a token weighting mechanism that prioritizes tokens with high informational density (estimated by intra-group information entropy), suppressing redundant tokens like generic reasoning prefixes. Furthermore, we reformulate RL training by shifting from single-choice to multi-choice QA tasks, where soft rewards enable finer-grained gradient estimation by distinguishing partial correctness. Additionally, we propose question-answer inversion, a data augmentation strategy to generate diverse multi-choice samples from existing benchmarks. Experiments demonstrate state-of-the-art performance on several video reasoning and general understanding benchmarks. Notably, TW-GRPO achieves 50.4\% accuracy on CLEVRER (18.8\% improvement over Video-R1) and 65.8\% on MMVU. Our codes are available at \href{https://github.com/longmalongma/TW-GRPO}{https://github.com/longmalongma/TW-GRPO}.

\end{abstract}

%% file: 1_intro.tex
\begin{figure}[t]
  \centering
\includegraphics[width=\textwidth]{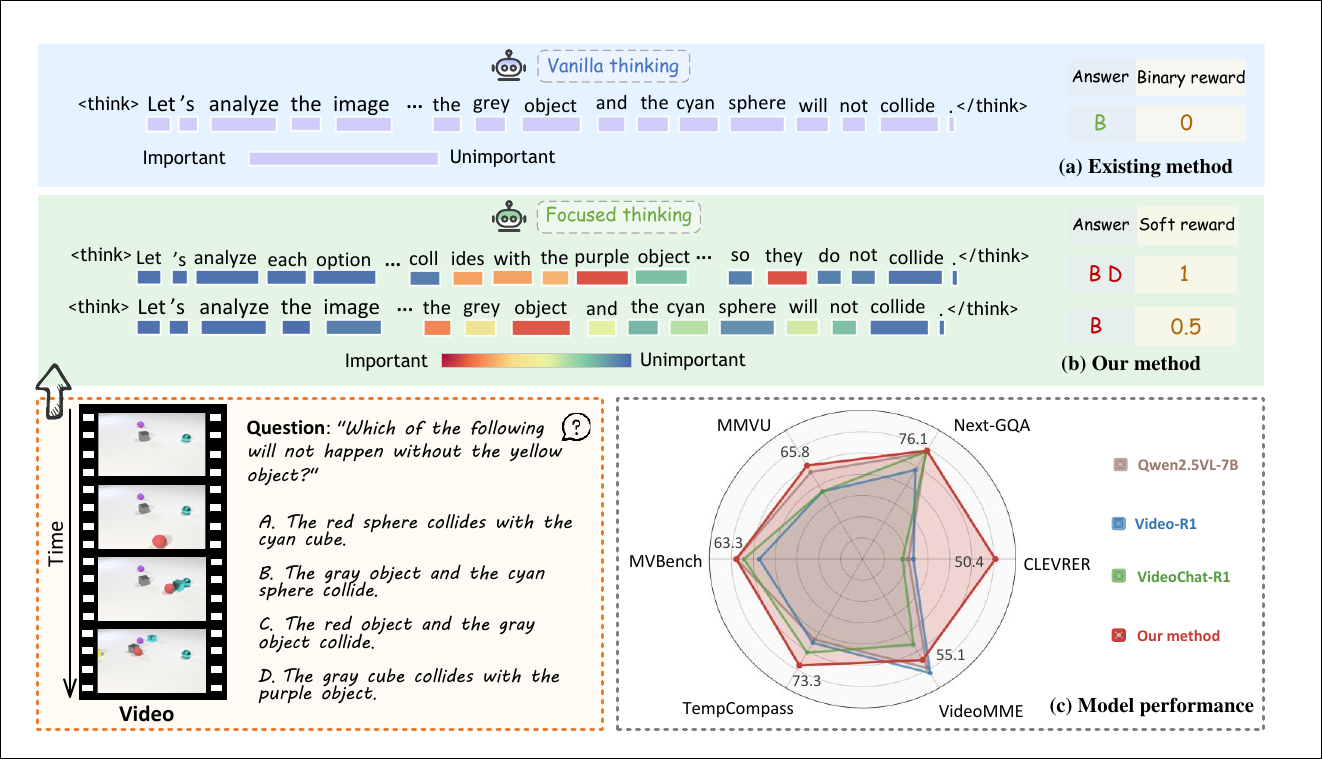} 
  \caption{
  TW-GRPO integrates focused thinking and soft multi-level rewards for multi-choice QA. Unlike vanilla thinking which assigns uniform token importance, focused thinking highlight critical tokens to dominate loss calculation. By shifting single-choice QA’s binary rewards to multi-choice QA’s multi-level rewards, TW-GRPO enables fine-grained gradient estimation and training efficiency.
  }    
  \vspace{-6mm}
  \label{fig:intro}
\end{figure}

\section{Introduction}

Recent advances in reinforcement learning (RL) for large language models (LLMs) have yielded significant improvements in reasoning capabilities. DeepSeek-R1~\cite{guo2025deepseek} demonstrated that pure RL optimization can substantially improve model reasoning, while subsequent works~\cite{chen2025r1v,huang2025vision,meng2025mm,zhou2025r1,peng2025lmm} extended these benefits to multimodal scenarios. Notable examples include VideoR1~\cite{videor1}, which introduced T-GRPO for video spatiotemporal reasoning, and VideoChat-R1~\cite{li2025videochat} that leverages GRPO-based multi-task joint fine-tuning. These improvements show promising progress in understanding of fine-grained video details and multi-step reasoning.

Although RL-based approaches excel in optimizing verifiable metrics, critical challenges persist in refining reasoning quality and reward granularity for complex multimodal tasks. \textbf{First}, while chain-of-thought (CoT) reasoning has proven effective for solving language-based intricate problems~\cite{shao2024deepseekmath}, its application to MLLMs often results yields verbose, unfocused thinking chains (\eg, overthinking~\cite{jiang2025mme}). Building upon this, current training objectives fail to prioritize semantically critical spatio-temporal cues, which may obscure pivotal information and hurt learning efficiency. \textbf{Second}, existing methods rely on sparse, binary rewards derived from single-choice question-answer (QA)  tasks~\cite{videor1,zhang2025r1,meng2025mm}. These rewards assign maximum credit for fully correct answers and none otherwise, disregarding partial correctness. Recent work on video grounding~\cite{li2025videochat}, shows that soft reward signals enable finer-grained optimization. However, such approaches remain underexplored in mainstream video QA tasks, where single-choice formats lack naturally defined multi-level reward signals.

To address these limitations, we propose TW-GRPO, a novel framework that enhances GRPO via token weighting and multi-grained rewards. As shown in Figure~\ref{fig:intro}, our dynamic weighting mechanism prioritizes tokens with high informational density during loss computation. 
Specifically, we estimate token importance by analyzing intra-group information entropy across token positions, focusing the model on content critical to reasoning outcomes rather than generic phrases (\eg, prefatory statements and repeated verifications). 
By prioritizing these tokens, the model learns to generate concise, task-aware reasoning chains while avoiding redundant or irrelevant details. 

Furthermore, we reformulate RL training using multi-choice QA tasks, replacing sparse binary rewards with multi-level ones. Our approach distinguishes between partially correct and fully incorrect answers, enabling finer-grained gradient estimation and stabilizes policy updates. To mitigate multi-choice data scarcity, we introduce Question-Answer Inverse (QAI), that converts single-choice tasks into multi-choice formats by negating questions and inverting answers.

Experiments demonstrate TW-GRPO’s superiority on multiple video reasoning and general understanding benchmarks. As shown in Table~\ref{tab:main_results}, our model achieves state-of-the-art accuracy on CLEVRER~\cite{yi2019clevrer}, NExT-GQA~\cite{0Can}, and MMVU~\cite{zhao2025mmvu}, outperforming Video-R1 with a clear margin by 18.8\%, 1.8\%, and 1.6\%, respectively. With focused thinking, qualitative analysis reveals condensed reasoning chains focused on critical visual or logical cues. Multi-level rewards also reduce reward variance during training. 
Our main contributions are summarized as follows:
\begin{itemize}

\item We propose dynamic token weighting, a mechanism prioritizing tokens with high informational density during loss computation, enabling concise, task-focused reasoning chains.

\item We propose multi-grained reward modeling using multi-choice QA tasks with partial correctness evaluation, improving gradient estimation and policy stability.

\item We propose question-answer inverse, a data augmentation converting single-choice QA into multi-choice formats via question negation and answer inversion, mitigating data scarcity.


\end{itemize}

%% file: 2-related-work.tex
\section{Related Works}

\subsection{Reinforcement Learning in MLLMs}
The reasoning abilities of LLMs have been a central focus of recent research, with efforts aimed at enhancing their capacity for complex, multi-step problem-solving tasks. RL has been a key driver of this progress, with works such as OpenAI-o1~\cite{jaech2024openai} and DeepSeek-R1~\cite{guo2025deepseek} achieving notable results. The latter adopts GRPO~\cite{shao2024deepseekmath}, a RL method that extends Proximal Policy Optimization (PPO)~\cite{zheng2023secrets} by eliminating the critic model and estimating relative quality through group-wise response comparisons, enabling efficient policy optimization. For MLLMs, numerous efforts~\cite{videor1,peng2025lmm, zhan2025visionr1, zhang2025r1, zhou2025r1,  li2025videochat, chen2025seed} have applied GRPO techniques with verifiable reward mechanisms to improve visual reasoning performance. However, existing GRPO-based methods operate at the sequence level and lack mechanisms to distinguish informative tokens. And as shown in Figure~\ref{fig:intro}, generic phrases like ``Let's think...'' are unnecessary for optimizing the policy model. Ignoring this variation can lead to misaligned optimization signals, resulting in verbose or redundant reasoning. To address this issue, we propose a token-level extension of GRPO that models token importance, enabling the policy to focus on tokens with high informational entropy and improving reasoning quality.

\subsection{MLLMs for Video Understanding}

Video understanding is a crucial capability for MLLMs, enabling them to interpret and reason over dynamic visual content~\cite{cheng2024videollama, shu2023audio, yan2024visa, zhang2025towards, zhang2024llava}.  
Recent advancements have resulted in the development of MLLMs specifically designed to improve video understanding tasks. Recent models such as VideoLLaMA2~\cite{cheng2024videollama} improve video-language alignment through spatiotemporal modeling and multimodal integration. Building on the success of DeepSeek-R1~\cite{guo2025deepseek}, the GRPO algorithm~\cite{videor1,li2025videochat} has been widely adopted in video reasoning tasks. However, existing RL-based frameworks rely on sparse, binary reward signals~\cite{videor1}, offering no distinction between partially correct and entirely incorrect responses in video QA. Recently, VideoChat-R1~\cite{li2025videochat} introduced an IoU-based soft reward for video grounding, providing continuous feedback that has proven effective in improving learning stability and precision. Inspired by this, we explore the use of soft reward mechanisms in Video-QA. Specifically, we reformulate the RL objective as a multi-choice classification problem, enabling multi-level reward assignment and fine-grained policy optimization.

%% file: 4-method.tex
\section{Methodology}

\begin{figure}[t]
  \centering
  \includegraphics[width=1\textwidth]{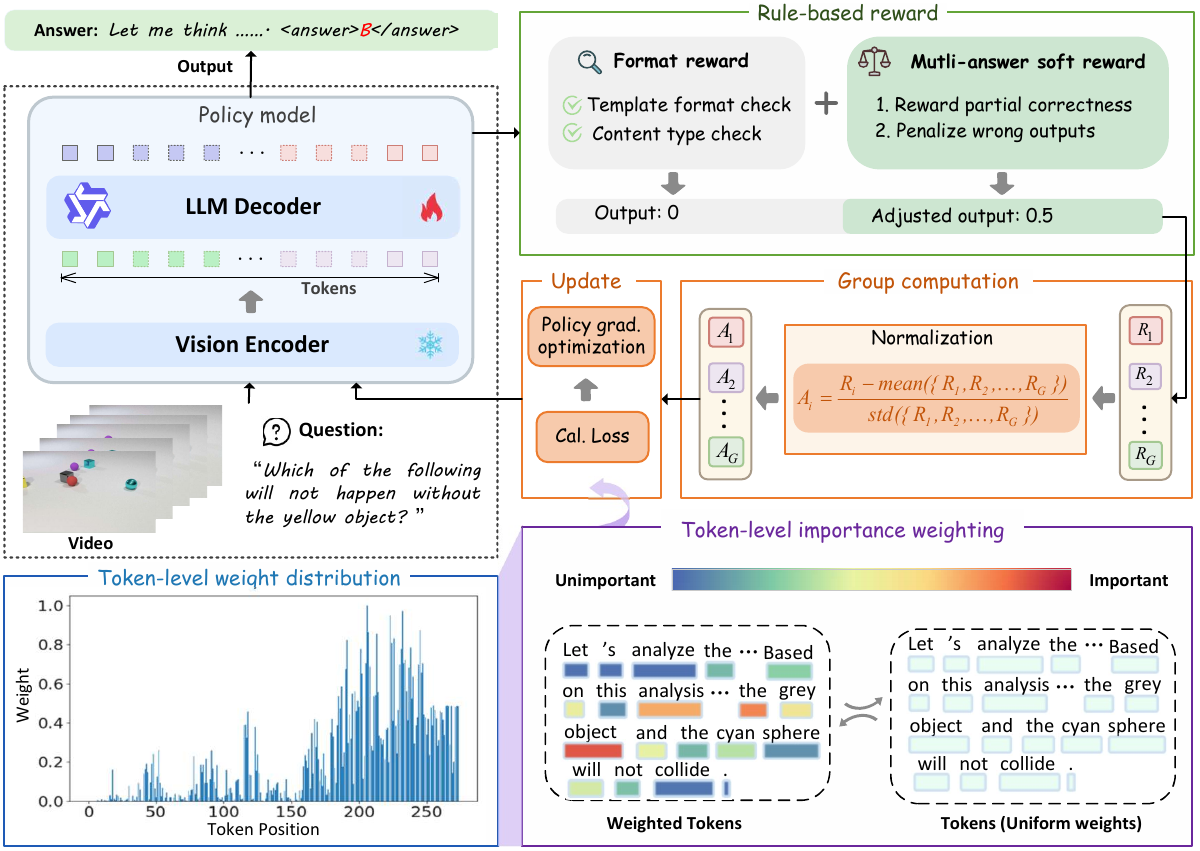} 
  \caption{Overview of the TW-GRPO framework. The diagram shows the key steps in a forward pass, starting from the video input, generating possible completions, and calculating the reward with adjustments for the final objective and model updates. Specifically, a multi-level soft reward is incorporated into the reward calculation, providing partial correctness feedback. These signals are then integrated into the final objective, where token-level importance weighting is applied, allowing the model to prioritize more informative tokens and improve overall performance.}
\vspace{-6mm}
  \label{fig:overview}
\end{figure}

\subsection{Preliminary}
\label{subsec:Preliminary}
As shown in Figure~\ref{fig:overview}, we focus on the task of multi-choice Video-QA, where the model is required to select the correct answer from a set of candidate options based on both the video content and the question. To enhance the performance of MLLMs on this task, one of the most advanced RL-based method, the GRPO~\cite{shao2024deepseekmath} has been introduced. For an input query $q$, GRPO samples $G$ candidate responses $o=\{o_1, \dots,o_G\}$ from the policy distribution. The rule-based reward model evaluates these responses to obtain reward scores $\{R_1, \dots, R_G\}$. The relative quality of each response is then computed through standardization:
\begin{equation}
\label{eq:ro}
    \hat{A_i}=
    \frac{R_i-\mathrm{mean}(\{R_i\}_{i=1}^G)}{\mathrm{std}(\{R_i\}_{i=1}^G)} \text{,}
\end{equation}
where $\hat{A_i}$ denotes the normalized advantage of the $i$-th response within the group. The optimization objective combines response quality improvement with policy regularization:
\begin{equation}
\begin{aligned}
&\mathcal{J}_\text{GRPO}(\theta) = \mathbb{E}_{(q,a)\sim \mathcal{D}, \{o_i\}_{i=1}^G\sim \pi_{\theta_\text{old}}(\cdot\mid q)} \\&
\Bigg[ \frac{1}{G}\sum_{i=1}^{G} \frac{1}{|o_i|}\sum_{t=1}^{|o_i|} \Bigg( 
\min \Big( r_{i,t}(\theta) \hat{A}_{i,t},  
\ \text{clip} \Big( r_{i,t}(\theta), 1 - \varepsilon, 1 + \varepsilon \Big) \hat{A}_{i,t} \Big)
- \beta D_{\text{KL}}(\pi_{\theta} || \pi_{\text{ref}}) 
\Bigg) \Bigg],
\label{eq:grpoloss}
\end{aligned}
\end{equation}
where
\begin{equation}
    r_{i,t}(\theta)=\frac{\pi_{\theta}(o_{i,t} \mid q, o_{i,<t})}{\pi_{\theta_{\text{old}}}(o_{i,t} \mid q,o_{i,<t})}.
\label{reward_computation}
\end{equation}
Despite its efficiency, existing GRPO algorithms do not differentiate between token positions during optimization, as shown in Equation~\ref{eq:grpoloss}. This leads the model to expend unnecessary effort on uninformative tokens, such as generic phrases like ``Let's think...'', which are less useful for optimizing the policy compared to tokens that describe critical spatiotemporal cues. In addition, since QA tasks are typically formulated as single-choice classification problems, existing approaches often adopt binary reward signals. However, in video grounding tasks~\cite{li2025videochat}, such binary signals have been shown to be less effective than continuous reward. Nevertheless, how to incorporate multi-level rewards into single-choice QA tasks remains unexplored.

To address these two issues, we propose the \textbf{TW-GRPO} framework, as illustrated in Figure~\ref{fig:overview}. In order to address the overlooked variation in token importance, we introduce a \textit{token-level importance wighting} mechanism in Section~\ref{Token_level_wighting}, which guides the model to focus on informative tokens. Furthermore, to overcome the limitations of binary rewards in single-choice QA, we reformulate the task as a \textit{multi-answer} setting, where a question may have one or more correct options. Then, a crafted \textit{multi-level soft reward} is designed to provide partial credit, enabling finer-grained policy learning, which is introduced in Section~\ref{reward_design}.

\subsection{Token-Level Importance Wighting}
\label{Token_level_wighting}

In this section, we address how to model token-level importance, enabling the policy to distinguish informative tokens better and optimize more effectively. As usual, the fine-grained assessment of reasoning quality typically requires an auxiliary critic model~\cite{zhang2025r1}, which introduces additional parameters and undermines one of GRPO's main advantages. Inspired by recent work~\cite{bigelow2024forking, lin2024critical}, which demonstrates that key reasoning tokens can be identified based on token-level distributional differences, we propose a lightweight approach grounded in the concept of information entropy. The key insight is that token positions where candidate outputs exhibit higher divergence from the expected distribution are more likely to carry more information. This allows us to estimate token importance without introducing extra model components.

Specifically, we purpose the token importance weight $w_t$ to quantify the information content of each token position. In detail, the Kullback-Leibler (KL) divergence $D_{\text{KL}}$ measures the discrepancy between the probability distribution of the token at position $t$ in the candidate output $o_i$ and the expected distribution at the same position. For token position $t$, we calculate the divergence $D_t$ as:
\begin{equation}
D_t = \sum_{i=1}^{G} D_{\text{KL}}\left( p(o_{i,t}) \big\Vert \mathbb{E}[o_t] \right),
\end{equation}
where $G$ denotes the number of candidate outputs, and $\mathbb{E}[o_t]$ is the expected probability distribution at token position $t$, which computed by averaging the probability distributions of each candidate output, with the missing tokens filled in using the uniform distribution $\mathcal{U}(V)$ to account for variable sequence lengths. This filling process ensures that all token sequences, regardless of their length, contribute fairly to the divergence calculation, preventing bias towards longer sequences and capturing the meaningful information in shorter sequences. To ensure numerical stability and comparable importance scores across different positions, we normalize the divergence measurements using min-max normalization:
\begin{equation}
w_t = (1 + \alpha) \cdot \frac{D_t - D_{\min}}{D_{\max} - D_{\min}}.
\end{equation}
In this formulation, $\alpha$ is a hyperparameter that controls the scaling of token importance. To ensure comparability, the raw divergence scores are normalized, mapping them to a standard range while preserving their relative differences. Moreover, the addition of the constant offset $(1 + \alpha)$ guarantees that tokens with low divergence retain non-zero weights, thus preventing any position from being entirely ignored during training. Consequently, the resulting weights $w_t$ enable position-sensitive optimization by modulating the learning signal according to token-level informativeness.
\begin{equation}
    \begin{aligned}
    \mathcal{J}_{\text{TW-GRPO}}(\theta) 
    = \ &\mathbb{E}_{(q,a)\sim \mathcal{D}, \{\alpha_i\}_{i=1}^G \sim \pi_{\theta_\text{old}}(\cdot \mid q)} \\
    &\Bigg[ \frac{1}{\sum_{i=1}^G |\alpha_i|} 
    \sum_{i=1}^G \sum_{t=1}^{|\alpha_i|} 
    \min \Big( w_t \cdot r_{i,t}(\theta) \hat{A}_{i,t}, \ \text{clip} \big( r_{i,t}(\theta), 1 - \varepsilon, 1 + \varepsilon \big) \hat{A}_{i,t} \Big) \Bigg].
    \end{aligned}
    \label{eq:tw_grpo_loss}
\end{equation}
Here, we normalize the number of outputs $|o_i| $ to balance their contributions to the overall loss~\cite{yu2025dapo}. Notably, our method does not require additional evaluation models to assess each step of the model's output. It only requires simple distance calculations to guide the model in applying varying levels of attention to different reasoning positions. After addressing the first issue, we focus on the second challenge: the need for multi-level reward signals in QA tasks.

\subsection{Multi-Answer Soft Reward}
\label{reward_design}
In this section, we address the inefficiency of reward signals in QA tasks caused by the single-choice formulation. Our solution consists of two main steps. \textbf{First}, inspired by multi-choice formats in standardized testing, we reformulate the single-choice QA task as a textit{multi-answer} setting, where each question contains at least one correct answer, and possibly more. \textbf{Second}, leveraging this multi-choice question, we introduce a \textit{multi-level soft reward} that assigns partial credit based on the overlap between the predicted and ground-truth answers. This enables more informative reward feedback for reinforcement learning. We describe the details of each component below.

\paragraph{Redefine Multi-choice QA Task.} In Video-QA tasks, standard benchmarks such as NExT-GQA~\cite{0Can} are typically formulated as single-choice questions, where each question has exactly one correct answer. As a result, evaluation is based on 0/1 accuracy, which inherently limits the ability to generate multi-level soft rewards. Interestingly, multi-answer questions, which are often used in the most challenging sections of standardized exams, align well with the needs. These questions offer both suitable difficulty and the presence of at least one correct option, making them well-suited to our training objective. Therefore, we reformulate the single-choice QA task as a multi-answer one. Unfortunately, this raises a new challenge: obtaining suitable multi-answer data for training. To the best of our knowledge, only a limited number of existing datasets, such as the counterfactual reasoning task in CLEVRER~\cite{yi2019clevrer}, address multi-choice problems.

To mitigate this scarcity, we introduce question-answer inversion, a novel data augmentation technique that transforms single-choice questions into multi-answer questions in general datasets. For example, in the NExT-GQA~\cite{0Can} dataset, a question such as “Why did the boy pick up one present from the group and move to the sofa?” is modified by changing "did" to "didn't," thereby transforming it from a five-choice, single-answer question into a five-choice, multiple-answer question. To prevent the model from incorrectly associating negation with the selection of multiple correct answers, we introduce a mechanism that randomly removes correct options, ensuring that the model remains challenged and is required to reason more carefully. Finally, by performing random question-answer inversion on the original dataset, we construct the final multi-choice NExT-GQA dataset, which may contain more than one correct answer, thereby increasing task complexity and fostering more sophisticated reasoning and deeper understanding of the underlying context.

However, while introducing multi-choice questions improves the training environment for RL, it also presents significant challenges. The increased complexity causes traditional accuracy reward mechanisms, based on binary accuracy (0 or 1), to exhibit significant reward variance between single-choice and multi-choice questions. As shown in Figure~\ref{fig:std_length}, the GRPO model trained on the single-choice dataset (GRPO(single-choice)) exhibits a notably lower reward standard deviation compared to the GRPO model trained on the multi-choice dataset (GRPO(fixed reward)). This higher variance complicates model convergence, making stable improvements in reasoning ability more difficult. Therefore, a key challenge is optimizing the fixed reward mechanism to mitigate this variance and ensure consistent progress in reasoning with the more complex multi-choice dataset.

\paragraph{Multi-Level Soft Reward.} To address this, we draw inspiration from the IoU reward used in grounding tasks such as VideoChat-R1 \cite{li2025videochat}, where a continuous-valued reward reflects the degree of temporal overlap between predicted and ground-truth intervals. This soft feedback enables the model to capture partial correctness and optimize both precision and recall, rather than relying solely on exact matches. Analogously, we proposed the multi-level soft reward,  which assigns graded credit to partially correct predictions and penalises completely incorrect ones. The reward is defined as:
\begin{equation}
R_{\mathrm{soft}} = 
\begin{cases}
\frac{|P|}{|G|}, & \text{if } P \subseteq G, \\
0, & \text{if } P \not\subseteq G.
\end{cases}
\label{eq:soft_accuracy}
\end{equation}
Here, $P$ denotes the predicted set and $G$ the ground truth set. If $P \subseteq G$, the reward is $|P|/|G|$; otherwise, it is 0. As shown in Figure~\ref{fig:intro}, if the ground truth is $\{B, D\}$ and the model predicts $\{B\}$, the reward is $1/2$, reflecting partial correctness. Predictions including elements outside $G$ receive a reward of 0, penalizing false positives. This design ensures that the model is proportionally rewarded for partially correct answers,  improving fine-grained gradient estimation and policy stability.

%% file: 5-experiment.tex
\section{Experiment}
\subsection{Experiment Setup}

We train our model based on Qwen2.5-VL-7B using two NVIDIA H800 GPUs with a lightweight setup of 500 RL steps on 1,000 CLEVRER counterfactual train datasets. Each frame is processed at a resolution of $128 \times 28 \times 28$ during training. For reasoning, the frame resolution is increased to $256 \times 28 \times 28$ while maintaining a maximum of 16 frames to improve performance. Evaluation is conducted on six video benchmarks covering general understanding and reasoning: MVBench~\cite{li2024mvbench}, TempCompass~\cite{liu2024tempcompass}, VideoMME~\cite{fu2024video}, MMVU~\cite{zhao2025mmvu}, NExT-GQA~\cite{0Can}, and CLEVRER~\cite{yi2019clevrer}. Detailed settings for all experiments are provided in Appendix~\ref{sec:detailed_setup}.

\begin{table*}[htpb]
\vspace{-2mm}
\label{combined_table}
\caption{Comparison of model performance on both video reasoning and general video benchmarks.}
\resizebox{\linewidth}{!}{%
\setlength{\tabcolsep}{0.8mm}
\renewcommand\arraystretch{1.3}
\begin{tabular}{@{}lcccc|ccc@{}}
\toprule
\multirow{2}{*}{Models} & \multirow{2}{*}{Training} & \multicolumn{3}{c}{Video Reasoning Benchmark} & \multicolumn{3}{c}{Video General Benchmark} \\ 
\cmidrule(l){3-8} 
 &  & CLEVRER$\rm_{cf}$ & NExT-GQA & MMVU$\rm _{mc}$ & MVBench & TempCompass & VideoMME$\rm _{(wo \ sub)}$ \\ 
\midrule
\rowcolor{gray!5}\multicolumn{8}{l}{\textbf{\textit{Baseline}}} \\
LLaMA-VID \cite{li2024llama}      & -  & -     & -     & -        & 41.9& 45.6    & -              \\
VideoLLaMA2 \cite{cheng2024videollama}    & -  & -     & -     & 44.8     & 54.6& -       & 47.9           \\
LongVA-7B \cite{zhang2024long}       & -  & -  & -  & -        & -   & 56.9    & 52.6           \\
Video-UTR-7B \cite{yu2025unhackable}   & -  & -     & -     & -        & 58.8& 59.7    & 52.6           \\
Kangeroo-8B \cite{liu2024kangaroo}    & -  & -     & -     & -        & 61.1& 62.5    & 56.0           \\ 
Qwen2.5-VL-7B (Zero-Shot)~\cite{bai2025qwen2} & - & 30.5 & 75.9 & 65.4 & \textbf{63.3} & 72.5 & 56.5 \\
Qwen2.5-VL-7B (CoT)~\cite{videor1} & - & 27.7 & 73.4 & 63.0 & 57.4 & 72.2 & 53.1 \\
\midrule
\rowcolor{gray!5}\multicolumn{8}{l}{\textbf{\textit{Supervised Finetuning}}} \\
Qwen2.5-VL-7B (SFT)~\cite{videor1} & 165K SFT & 29.0 & 69.0 & 61.3 & 59.4 & 69.2 & 52.8 \\
\midrule
\rowcolor{gray!5}\multicolumn{8}{l}{\textbf{\textit{Reinforcement Learning Finetuned}}} \\
Video-R1~\cite{videor1} & 165K SFT + 4K RL & 31.6 & 74.3 & 64.2 & 62.7 & 72.6 & \textbf{57.4} \\
VideoChat-R1~\cite{li2025videochat} & 18K RL & 29.2 & 76.0 & 64.2 & 63.1 & 72.9 & 52.4 \\
{GRPO} & 1K RL & {41.1} & {75.2} & {65.1} & {62.8} & {71.9} & 55.9 \\
\rowcolor{gray!15}
\textbf{TW-GRPO (Ours)} & 1K RL & \textbf{50.4} & \textbf{76.1} & \textbf{65.8} & \textbf{63.3} & \textbf{73.3} & 55.1 \\
\toprule
\end{tabular}
}
\label{tab:main_results}
\vspace{-2mm}
\end{table*}

\begin{figure*}[t]
    \centering
    \includegraphics[width=1\textwidth]{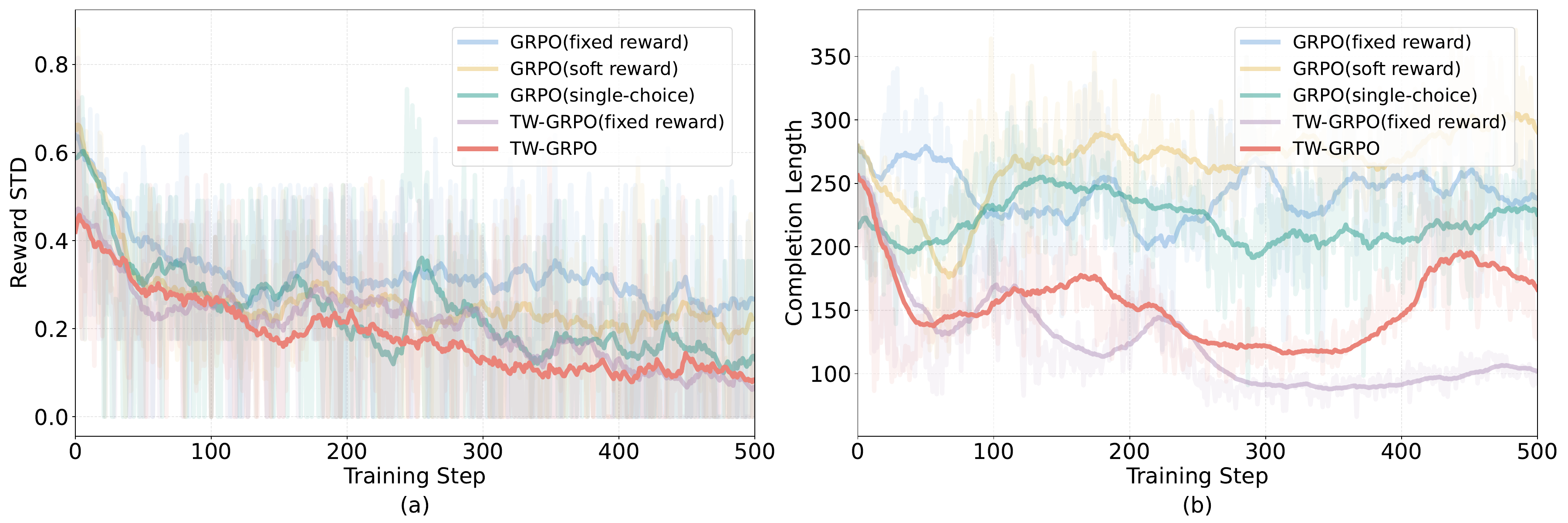}
    \caption{\textbf{Training dynamics of different GRPO variants.} 
    (a) TW-GRPO achieves faster convergence in reward standard deviation, indicating more stable and efficient learning.
    (b) It also produces consistently shorter output lengths, reflecting more concise and effective reasoning than other methods.}
    \label{fig:std_length}
    \vspace{-6mm}
\end{figure*}

\subsection{Main Results}


\paragraph{Superior Performance of TW-GRPO.}
As shown in Table~\ref{tab:main_results}, TW-GRPO consistently outperforms existing models in both video reasoning and general understanding tasks, achieving better results with fewer training samples. Specifically, in reasoning tasks such as CLEVRER, NExT-GQA, and MMVU, TW-GRPO demonstrates significant improvement over the original GRPO models that do not utilize soft rewards and token-level weighting. On CLEVRER, it reaches 50.4\% accuracy, surpassing the next best (Video-R1) by over 18\%. It also beats Video-R1 and VideoChat-R1 on NExT-GQA and MMVU by 1.8\% and 1.6\%, respectively. For general video understanding tasks, TW-GRPO demonstrates competitive performance, even with fewer training resources. In MVBench, TW-GRPO matches the zero-shot performance of Qwen2.5-VL-7B (63.3\%), while outperforming both Video-R1 and VideoChat-R1. In TempCompass, TW-GRPO leads with a 73.3\% accuracy, surpassing the best-performing baseline by 0.4\%. Though its performance in VideoMME is slightly lower, TW-GRPO still outperforms VideoChat-R1 by 2.7\%. Even under identical training conditions, TW-GRPO significantly improves over GRPO across five benchmarks. These results highlight the effectiveness and robustness of our method. By enhancing reinforcement learning with token-level importance weighting and multi-level reward strategies, it provides more efficient and stable policy learning, thereby boosting model performance across diverse tasks.

\paragraph{Training Dynamics and Convergence Behavior.} Figure~\ref{fig:std_length} illustrates the training dynamics of different GRPO variants. Figure~\ref{fig:std_length}(a) shows that TW-GRPO achieves faster convergence in reward standard deviation, indicating more stable learning. This stability is attributed to the introduction of multi-level soft reward and token-weighting strategies, which help the model handle ambiguous questions more effectively. In detail, traditional GRPO suffers from slow convergence on multi-choice tasks due to fixed accuracy rewards. In contrast, our soft reward strategy reduces the reward standard deviation, leading to more stable optimization. Furthermore, the token-level importance weighting mechanism, which helps the model focus on more informative tokens, thereby improving optimization efficiency and speeding up convergence. As shown in Figure~\ref{fig:std_length}(b), TW-GRPO produces shorter output sequences, suggesting it learns to reason more concisely. This reflects a more substantial alignment between the reward objective and the final model behaviour, confirming the effectiveness of our proposed training design. Due to this crafted design, TW-GRPO achieves smoother convergence with fewer tokens in the generated outputs, reflecting more concise reasoning.

\begin{figure}[t]
  \centering
  \includegraphics[width=1\textwidth]{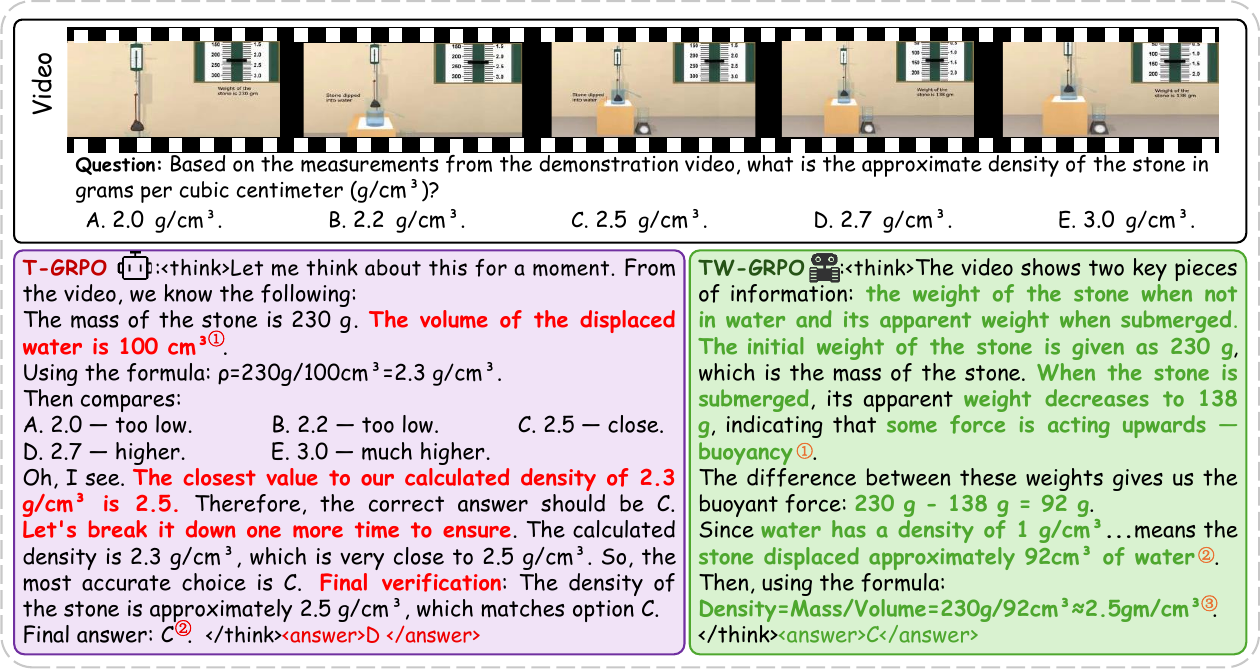} 
  \caption{Comparison of reasoning paths from T-GRPO and TW-GRPO on MMVU samples.}
  \label{fig:mmvu_sample}
\vspace{-6mm}
\end{figure}

\paragraph{Qualitative Analysis of Reasoning Path.} We compare T-GRPO and TW-GRPO on a physics-based density estimation task from the MMVU dataset. As shown in Figure~\ref{fig:mmvu_sample}, a stone is first weighed in air (230\,g) and then submerged in water, where its apparent weight drops to 138\,g. Solving the task requires applying Archimedes' principle to derive the displaced volume from the 92\,g buoyant force and compute the density as mass over volume. The T-GRPO model attempts this but incorrectly assumes a volume of 100\,cm$^3$ \circled[red]{1}, leading to a calculated density of 2.3\,g/cm$^3$. It then mistakenly claims 2.5\,g/cm$^3$ is the closest answer, despite 2.2\,g/cm$^3$ being numerically closer \circled[red]{2}. Although the model attempts to refine its performance through reflection, it remains anchored to an incorrect conclusion, resulting in substantial token inefficiency. Moreover, it ultimately selects a value of 2.7g/cm³, contradicting its prior (and redundant) estimate. In contrast, TW-GRPO trained model accurately extracts the key values from the video, applies physical principles to infer volume from buoyant force \circled[deeporange]{1}, and correctly matches the result to the provided answer choices \circled[deeporange]{2}\circled[deeporange]{3}. This example illustrates TW-GRPO's improved capacity for grounded, causal, and quantitative reasoning based on dynamic visual cues. Additionally, we also provide more visual samples in Appendix~\ref{visualization}.

\subsection{Ablation Study}

To better understand the contributions of each component in our method, we conducted ablation studies to assess the improvements brought by indeterminate selection and TW-GRPO. In addition to using accuracy as an evaluation metric, we compute soft accuracy based on Equation~\ref{eq:soft_accuracy}, enabling a more fine-grained assessment of the model's performance on multi-choice datasets.

We evaluate the necessity of adding uncertain options by assessing the performance of the training model on datasets with different problem types. Specifically, the multi-choice training data used for the STAR and NExT-GQA datasets is generated through our question-Answer inversion method. As shown in Table \ref{tab:data_construction}, for datasets such as CLEVRER, NExT-GQA, and STAR \cite{wu2024star}, multi-choice training outperforms training with only single-choice questions. The results demonstrate that incorporating multi-choice questions increases the complexity of the training set, thereby enhancing the model's reasoning ability. For instance, the accuracy of the model trained with multi-choice questions on the CLEVRER dataset is 41.1\%, which is 8.5\% higher than that of the GRPO model trained with only single-choice questions. The counterfactual reasoning subset of the CLEVRER dataset~\cite{yi2019clevrer} includes samples involving object dynamics and hypothetical outcome predictions. This provides a more complex and realistic training environment for models..

Next, we investigate the effectiveness of TW-GRPO under different training setups, reward strategies, and token weighting configurations, as summarized in Table~\ref{tab:final_ablation_full}. We first observe that training on multi-choice question types yields better performance than single-choice only settings, especially on CLEVRER, where multi-choice training improves accuracy by over 10\% compared to single-choice training. Regarding reward design, soft rewards consistently outperform fixed rewards across settings, improving soft accuracy by up to 8.7\% in multi-choice tasks. This indicates that multi-answer soft reward provides more informative and stable supervision signals, especially for ambiguous or partially correct answers. Finally, we assess the effect of token weighting. In both fixed and soft reward settings, TW-GRPO shows notable gains over GRPO. For example, with soft reward, TW-GRPO improves multiple-choice soft accuracy from 57.6\% to 64.4\%. When token weights are removed (TW-GRPO) 
 ($\alpha=0$)), performance drops, confirming that token-level importance modeled is crucial for fine-grained optimization. And the analysis of hyperparameters is provided in Appendix~\ref{ass_exp}.

\begin{table*}[t]
\small
\centering
\caption{Ablation study on data construction with different sources and sampling strategies.
We evaluate the effect of using single-choice and multi-choice questions from CLEVRER, NExT-GQA, and STAR datasets.}
\label{tab:data_construction}
\renewcommand{\arraystretch}{1.2}
\setlength{\tabcolsep}{1.5mm}
\begin{tabular}{l|c|cc|cc|c}
\toprule
\textbf{Setting} & \textbf{Single.} & \multicolumn{2}{c|}{\textbf{Multiple Choice}} & \multicolumn{2}{c|}{\textbf{All}} & \textbf{MMVU} \\
& Acc. (\%) & Acc. (\%) & Soft. (\%) & Acc. (\%) & Soft. (\%) & Acc. (\%) \\
\midrule
\rowcolor{gray!5}\multicolumn{7}{l}{\textit{\textbf{CLEVRER$\rm_{cf}$ Source}}} \\
GRPO (single) & \textbf{51.5} & 18.2 & 50.9 & 32.6 & 51.2 & 62.7 \\
\rowcolor{gray!5}\rowcolor{gray!5} GRPO (multi-choice) & 50.4 & \textbf{32.3} & \textbf{55.4} & \textbf{41.1} & \textbf{52.7} & \textbf{65.1} \\
\midrule
\rowcolor{gray!5}\multicolumn{7}{l}{\textit{\textbf{NExT-GQA Source}}} \\
GRPO (single-choice) & 48.0 & \textbf{12.3} & \textbf{45.5} & 30.7 & 45.7 & 64.3 \\
\rowcolor{gray!5} GRPO (multi-choice) & \textbf{63.5} & 9.6 & 44.9 & \textbf{36.7} & \textbf{54.7} & \textbf{64.6} \\
\midrule
\rowcolor{gray!5}\multicolumn{7}{l}{\textit{\textbf{STAR Source}}} \\
GRPO (single-choice) & 65.1 & \textbf{2.6} & \textbf{41.4} & \textbf{29.3} & \textbf{51.5} & 64.8 \\
\rowcolor{gray!5} GRPO (multi-choice) & \textbf{65.9} & 1.5 & 40.8 & 29.1 & \textbf{51.5} & \textbf{66.2} \\
\bottomrule
\end{tabular}
\vspace{-4mm}
\end{table*}

\begin{table*}[t]
\small
\centering
\caption{Ablation study on training type, reward design, and the token-level weighting mechanism (TW-GRPO). We evaluate performance on CLEVRER and MMVU under different settings.}
\label{tab:final_ablation_full}
\renewcommand{\arraystretch}{1.15}
\setlength{\tabcolsep}{1.5mm}
\begin{tabular}{l|c|cc|cc|c}
\toprule
\textbf{Setting} & \textbf{Single.} & \multicolumn{2}{c|}{\textbf{Multiple Choice}} & \multicolumn{2}{c|}{\textbf{All}} & \textbf{MMVU} \\
& Acc. (\%) & Acc. (\%) & Soft. (\%) & Acc. (\%) & Soft. (\%) & Acc. (\%) \\
\midrule
\rowcolor{gray!5}\multicolumn{7}{l}{\textit{\textbf{Training Problem Type (TW-GRPO)}}} \\
Single-choice  & 60.4 & 22.8 & 55.9 & 38.9 & 57.8 & 63.7 \\
\rowcolor{gray!5}Multi-choice                & \textbf{60.9} & 42.5 & 64.4 & \textbf{50.4} & \textbf{62.9} & \textbf{65.8} \\
\midrule
\rowcolor{gray!5}\multicolumn{7}{l}{\textit{\textbf{Reward Design (TW-GRPO, multi-choice)}}} \\
Fixed reward & \textbf{64.9} & 26.0 & 54.2 & 42.6 & 58.7 & 65.0 \\
\rowcolor{gray!5}Soft reward  & 60.9 & \textbf{42.5} & \textbf{64.4} & \textbf{50.4} & \textbf{62.9} & \textbf{65.8} \\
\midrule
\rowcolor{gray!5}\multicolumn{7}{l}{\textit{\textbf{Effect of Token Weighting (multi-choice, Fixed Reward)}}} \\
GRPO& 50.4 & \textbf{32.3} & 55.4 & 41.1 & 52.7 & \textbf{65.1} \\
\rowcolor{gray!5} TW-GRPO& \textbf{64.9} & 26.0 & \textbf{54.2} & \textbf{42.6} & \textbf{58.7} & 65.0 \\
\midrule
\rowcolor{gray!5}\multicolumn{7}{l}{\textit{\textbf{Effect of Token Weighting (multi-choice, Soft Reward)}}} \\
GRPO& 58.3 & 28.1 & 57.6 & 41.2 & 57.9 & 64.6 \\
TW-GRPO ($\alpha=0$)  & \textbf{64.7} & {36.6} & {60.5} & {48.6} & {62.3} & {62.1} \\
\rowcolor{gray!5} TW-GRPO& {60.9} & \textbf{42.5} & \textbf{64.4} & \textbf{50.4} & \textbf{62.9} & \textbf{65.8} \\
\bottomrule
\end{tabular}
\vspace{-4mm}
\end{table*}

%% file: 6-conclusion.tex
\section{Conclusions}

In this work, we present TW-GRPO, a novel reinforcement learning framework that advances video reasoning in MLLMs. While prior approaches have improved model accuracy, two core limitations remain: the inability to distinguish token-level contributions and the inefficiency of binary reward signals. TW-GRPO addresses these challenges by incorporating token-level importance weighting and introducing multi-answer soft rewards that grant partial credit for partially correct responses. Extensive experiments across six benchmarks, along with comprehensive ablation studies, validate the effectiveness of our approach. We hope this insight provides a foundation for future research in fine-grained video reasoning with MLLMs.

%% file: 7-supplementary.tex
\newpage

\section{Discussion on Entropy-based Measurement}

We note that some concurrent studies~\cite{agarwal2025unreasonable, chen2025seed, zhang2025right, zhao2025learning} have consistently highlighted the value of entropy-based signals as intrinsic measures that significantly enhance policy optimization in LLM reasoning. Similarly, our method also integrates entropy into the optimization process, but introduces a novel mechanism centered on token-level informativeness through distributional divergence.

\paragraph{Entropy Minimization in Reasoning Optimization.} Works such as EMPO~\cite{zhang2025right} and SEED-GRPO~\cite{chen2025seed} adopt a semantic perspective, where entropy is computed over clusters of sampled completions. EMPO minimizes entropy among latent semantic groups formed from multiple outputs, encouraging global consistency across generations. SEED-GRPO builds upon this by adjusting GRPO updates based on the entropy level of input prompts—assigning larger updates to low-entropy (confident) inputs and smaller updates to uncertain ones. These approaches frame entropy as a measure of reasoning confidence at the sequence level.

\paragraph{Token-Level Entropy and Local Uncertainty.} While EMPO and SEED-GRPO focus on semantic-level aggregation, recent work has revisited the benefits of entropy minimization at the token level. Agarwal et al.~\cite{agarwal2025unreasonable} introduced EM-RL and EM-FT as two forms of token-level entropy-based optimization. In EM-RL, token-level entropy is used as the sole reward signal in reinforcement learning, promoting deterministic generation at each step. Meanwhile, EM-FT applies direct token-level entropy minimization as a fine-tuning objective, reinforcing confidence locally across generation trajectories. Importantly, these approaches highlight that entropy minimization can be interpreted as a mechanism to exploit pretrained confidence priors embedded in LLMs.

\paragraph{Self-Certainty and Internal Feedback.} Complementary to entropy-based reward signals, Zhao et al.~\cite{zhao2025learning} proposed INTUITOR, which replaces verifiable external rewards in GRPO with an intrinsic measure called self-certainty~\cite{kang2025scalable}, formulated as the KL divergence between the model's output distribution and a uniform distribution at each token step. By encouraging high self-certainty scores, the model aligns itself to more confident and consistent generation patterns, improving both in-domain reasoning and out-of-domain generalization. This work reinforces the idea that token-level distributional sharpness, much like entropy or self-certainty, can serve as a rich intrinsic feedback signal in reasoning-centric LLM optimization.

\paragraph{Our Approach: Informativeness via Distributional Divergence.} Similar to the above works, our method also incorporates entropy-based signals into the GRPO optimization process. However, we adopt a distinct perspective by modeling token-level informativeness through distributional divergence. Specifically, we compute the Kullback-Leibler (KL) divergence between the predicted distribution at each token position and the expected distribution aggregated across multiple candidate outputs. This divergence reflects the variability or uncertainty associated with each position, and serves as a proxy for its relative importance in the reasoning process. This approach aligns with the intuition behind self-certainty~\cite{zhao2025learning,kang2025scalable}, where confident and consistent predictions are indicative of stronger reasoning. In our case, token positions with low divergence typically exhibit stable distributions across samples, suggesting that the model has already formed confident decisions at these positions. Further optimization of such low-information tokens is often redundant and provides limited gains. In contrast, tokens with high divergence correspond to areas of uncertainty or critical decision points that are more likely to influence the overall output. By assigning higher weights to these informative positions, our method selectively amplifies learning signals where they are most needed, resulting in more effective and interpretable optimization within the GRPO framework.

\section{Details of TW-GRPO}

\subsection{Details of Optimization Objective}
We begin by reviewing the Group Relative Policy Optimization (GRPO) framework~\cite{shao2024deepseekmath}. Given an input query \( q \), GRPO samples \( G \) candidate responses \( o = \{o_1, \dots, o_G\} \) from the policy distribution \( \pi_{\theta_{\text{old}}} \). A rule-based reward model then assigns scalar reward scores \( \{R_1, \dots, R_G\} \) to these responses.

To quantify the relative quality of each response, the rewards are standardized within the group:
\begin{equation}
\label{eq:ro_ap}
    \hat{A_i}=
    \frac{R_i-\mathrm{mean}(\{R_i\}_{i=1}^G)}{\mathrm{std}(\{R_i\}_{i=1}^G)} \text{,}
\end{equation}
where $\hat{A_i}$ denotes the normalized advantage of the $i$-th response within the group. The GRPO objective encourages response quality improvements while regularizing the policy through a KL-divergence term:
\begin{equation}
\begin{aligned}
&\mathcal{J}_\text{GRPO}(\theta) = \mathbb{E}_{(q,a)\sim \mathcal{D}, \{o_i\}_{i=1}^G\sim \pi_{\theta_\text{old}}(\cdot\mid q)} \\&
\Bigg[ \frac{1}{G}\sum_{i=1}^{G} \frac{1}{|o_i|}\sum_{t=1}^{|o_i|} \Bigg( 
\min \Big( r_{i,t}(\theta) \hat{A}_{i,t},  
\ \text{clip} \Big( r_{i,t}(\theta), 1 - \varepsilon, 1 + \varepsilon \Big) \hat{A}_{i,t} \Big)
- \beta D_{\text{KL}}(\pi_{\theta} || \pi_{\text{ref}}) 
\Bigg) \Bigg],
\label{eq:grpoloss_ap}
\end{aligned}
\end{equation}
where
\begin{equation}
    r_{i,t}(\theta)=\frac{\pi_{\theta}(o_{i,t} \mid q, o_{i,<t})}{\pi_{\theta_{\text{old}}}(o_{i,t} \mid q,o_{i,<t})}.
\label{eq:r_i_ap}
\end{equation}

Although GRPO successfully eliminates the need for a critic model, as required in PPO~\cite{schulman2017proximal}, recent findings~\cite{yu2025dapo} reveal that its sample-level optimization and KL-divergence regularization may restrict the model's reasoning capacity, particularly in complex generation tasks. Motivated by these insights, we adopt a token-level policy gradient objective inspired by DAPO~\cite{yu2025dapo}, and remove the KL penalty to enable more flexible optimization dynamics. The resulting optimized objective is defined as:
\begin{equation}
    \begin{aligned}
    \mathcal{J}_{\text{GRPO}^{'}(\theta)} 
    = \ &\mathbb{E}_{(q,a)\sim \mathcal{D}, \{o_i\}_{i=1}^G \sim \pi_{\theta_\text{old}}(\cdot \mid q)} \\
    &{\Bigg[\frac{1}{\sum_{i=1}^G |o_i|} 
    \sum_{i=1}^G \sum_{t=1}^{|o_i|}} 
    \min \Big( r_{i,t}(\theta) \hat{A}_{i,t}, \ \text{clip} \big( r_{i,t}(\theta), 1 - \varepsilon, 1 + \varepsilon \big) \hat{A}_{i,t} \Big) \Bigg].
    \end{aligned}
    \label{eq:dapo_loss_ap}
\end{equation}

Building upon this token-level optimization objective, we now introduce our token-level importance modeling approach, which aims to explicitly identify and leverage the most influential tokens in the optimization process.

\subsection{Token-Level Importance Modeling}
\subsubsection{Theoretical Analysis and Motivation}
In the context of exploring multi-modal reward functions such as R1, recent implementations often follow the design choices in R1-V~\cite{chen2025r1v} and TRL~\cite{vonwerra2022trl}, which simplify the optimization formulation to improve training efficiency. Specifically,by assuming that the policy undergoes relatively small updates, i.e., $r_{i,t}(\theta)\in(1-\epsilon,1+\epsilon)$, the objective can be simplified as:
\begin{equation}
    \mathcal{J}_{\text{GRPO}^{'}(\theta)}= \mathbb{E}_{(q,a)\sim \mathcal{D}, \{o_i\}_{i=1}^G \sim \pi_{\theta_\text{old}}(\cdot \mid q)} {\Bigg[ \frac{1}{\sum_{i=1}^G |o_i|} 
    \sum_{i=1}^G \sum_{t=1}^{|o_i|}} 
     \Big( r_{i,t}(\theta) \hat{A}_{i,t} \Big) \Bigg],
    \label{eq:tw_grpo_sim}
\end{equation}
by padding all sampled responses $o_i$ to a uniform length $o_{\text{max}}=\mathop{\max}\limits_{i}|o_i|$ using a uniform distribution, and applying the commutative law of summation, we can derive from Eq.~\ref{eq:tw_grpo_sim} that:
\begin{equation}
    \mathcal{J}_{\text{GRPO}^{'}(\theta)}= \mathbb{E}_{(q,a)\sim \mathcal{D}, \{o_i\}_{i=1}^G \sim \pi_{\theta_\text{old}}(\cdot \mid q)} {\Bigg[ \frac{1}{\sum_{i=1}^G o_{\max}} 
    \sum_{t=1}^{o_{\max}} \sum_{i=1}^G } 
     \Big( r_{i,t}(\theta) \hat{A}_{i,t} \Big) \Bigg].
    \label{eq:tw_grpo_sim1}
\end{equation}

Since the adopted reward model operates at the sample level, it assigns rewards based solely on the complete sampled response and is inherently independent of individual token positions. As a result, the advantage computed in Eq.~\ref{eq:ro_ap} is also position-invariant for token index $t$, i.e., the following holds:
\begin{equation}
    \hat{A}_{i}=\hat{A}_{i,t},
    \label{eq:a_i}
\end{equation}
so, the Eq.~\ref{eq:tw_grpo_sim} can be expressed as
\begin{equation}
    \mathcal{J}_{\text{GRPO}^{'}(\theta)}= \mathbb{E}_{(q,a)\sim \mathcal{D}, \{o_i\}_{i=1}^G \sim \pi_{\theta_\text{old}}(\cdot \mid q)} {\Bigg[ \frac{1}{\sum_{i=1}^G o_{\max}} 
    \sum_{t=1}^{o_{\max}} \sum_{i=1}^G } 
     \Big( r_{i,t}(\theta) \hat{A}_{i} \Big) \Bigg].
    \label{eq:tw_grpo_sim_ai}
\end{equation}
when we define $\bar{r}_t \triangleq{\text{mean}(\{r_{i,t}\}^G_i)}$, Eq.~\ref{eq:tw_grpo_sim_ai} is equal to:
\begin{equation}
    \mathcal{J}_{\text{GRPO}^{'}(\theta)}= \mathbb{E}_{(q,a)\sim \mathcal{D}, \{o_i\}_{i=1}^G \sim \pi_{\theta_\text{old}}(\cdot \mid q)} {\Bigg[ \frac{1}{\sum_{i=1}^G o_{\max}} 
    \sum_{t=1}^{o_{\max}} \sum_{i=1}^G } 
     \Big( [(r_{i,t}(\theta)-\bar{r}_t)+\bar{r}_t] \hat{A}_{i} \Big) \Bigg].
    \label{eq:tw_grpo_sim_ai1}
\end{equation}
This can be separated into two terms:
\begin{equation}
\begin{aligned}
\mathcal{J}_{\text{GRPO}^{'}(\theta)}= &\mathbb{E}_{(q,a)\sim \mathcal{D}, \{o_i\}_{i=1}^G \sim \pi_{\theta_\text{old}}(\cdot \mid q)}\\
&\Bigg[ \frac{1}{\sum_{i=1}^G o_{\max}}
\sum_{t=1}^{o_{\max}} \sum_{i=1}^G 
(r_{i,t}(\theta) - \bar{r}_t) \hat{A}_i + \frac{1}{\sum_{i=1}^G o_{\max}} 
\sum_{t=1}^{o_{\max}} \sum_{i=1}^G 
\bar{r}_t \hat{A}_i \Bigg].
\end{aligned}
\label{eq:grpo_split}
\end{equation}

Due to the group-wise normalization of advantages (cf. Eq.~\ref{eq:ro_ap}), we have:
\begin{equation}
\sum_{i=1}^G \hat{A}_i = 0,
\label{eq:adv_sum_zero}
\end{equation}
which causes the second term in Eq.~\ref{eq:grpo_split} to vanish. Hence, the objective simplifies to:
\begin{equation}
\mathcal{J}_{\text{GRPO}'(\theta)} = 
\mathbb{E}_{(q,a)\sim \mathcal{D}, \{o_i\}_{i=1}^G \sim \pi_{\theta_\text{old}}(\cdot \mid q)} \left[ \frac{1}{\sum_{i=1}^G o_{\max}} 
\sum_{t=1}^{o_{\max}} \sum_{i=1}^G 
(r_{i,t}(\theta) - \bar{r}_t) \hat{A}_i \right].
\label{eq:grpo_reduced}
\end{equation}

To simplify the expression of $r_{i,t}$, we aim to approximate its denominator in a way that is both computationally efficient and statistically stable. To justify this approximation, we make the following assumption:

\begin{assumption}
\label{assumption}
We assume that (i) the number of sampled trajectories $G$ is sufficiently large, and (ii) the policy $\pi_{\theta}$ produces relatively stable outputs across similar histories. That is, for a fixed input $q$, the conditional distributions $\pi_{\theta}(\cdot \mid q, o_{j,<t})$ exhibit only mild variation across different $j$.
\end{assumption}

As empirically verified in Section~\ref{ass_exp}, this assumption is foundational for our method to achieve strong performance. Under this assumption, the individual conditional likelihoods $\pi_{\theta_{\text{old}}}(o_{i,t} \mid q, o_{i,<t})$ can be reasonably approximated by their average over the sample set. Specifically, we define the empirical distribution under the old policy at position $t$ as:
\begin{equation}
\pi^{\text{emp}}_{\theta_{\text{old}}, t} := \frac{1}{G} \sum_{j=1}^G \pi_{\theta_{\text{old}}}(o_{j,t} \mid q, o_{j,<t}).
\label{eq:emp_old}
\end{equation}
We then approximate the denominator of $r_{i,t}$ using this empirical estimate:
\begin{equation}
r_{i,t}(\theta) \approx \frac{\pi_{\theta}(o_{i,t} \mid q, o_{i,<t})}{\pi^{\text{emp}}_{\theta_{\text{old}}, t}}.
\end{equation}

Similarly, the mean importance ratio can be approximated as:
\begin{equation}
\bar{r}_t \approx \frac{1}{G} \sum_{j=1}^G \frac{\pi_{\theta}(o_{j,t} \mid q, o_{j,<t})}{\pi^{\text{emp}}_{\theta_{\text{old}}, t}} = \frac{\pi^{\text{emp}}_{\theta, t}}{\pi^{\text{emp}}_{\theta_{\text{old}}, t}},
\end{equation}
where we define:
\begin{equation}
\pi^{\text{emp}}_{\theta, t} := \frac{1}{G} \sum_{j=1}^G \pi_{\theta}(o_{j,t} \mid q, o_{j,<t}),
\label{eq:emp_new}
\end{equation}
as the empirical distribution of the current policy at position $t$. Substituting into the difference $(r_{i,t} - \bar{r}_t)$, we obtain:
\begin{equation}
r_{i,t} - \bar{r}_t \approx \frac{1}{\pi^{\text{emp}}_{\theta_{\text{old}}, t}} \left( \pi_{\theta}(o_{i,t} \mid q, o_{i,<t}) - \pi^{\text{emp}}_{\theta, t} \right).
\end{equation}

Substituting the approximation into Eq.~\ref{eq:grpo_reduced}, we obtain the final reformulated objective:
\begin{equation}
\begin{aligned}
\mathcal{J}_{\text{GRPO}'{(\theta)}} &= 
\mathbb{E}_{(q,a)\sim \mathcal{D}, \{o_i\}_{i=1}^G \sim \pi_{\theta_\text{old}}(\cdot \mid q)} \\ &\left[ \frac{1}{\sum_{i=1}^G o_{\max}} 
\sum_{i=1}^G \sum_{t=1}^{o_{\max}}
\frac{1}{\pi^{\text{emp}}_{\theta_{\text{old}}, t}} 
\left( \pi_{\theta}(o_{i,t} \mid q, o_{i,<t}) - \pi^{\text{emp}}_{\theta, t} \right) \hat{A}_i \right],
\label{eq:grpo_final1}
\end{aligned}
\end{equation}
which is equivalent to a re-ordered form:
\begin{equation}
\begin{aligned}
\mathcal{J}_{\text{GRPO}'{(\theta)}} &= 
\mathbb{E}_{(q,a)\sim \mathcal{D}, \{o_i\}_{i=1}^G \sim \pi_{\theta_\text{old}}(\cdot \mid q)} \\ &\left[ \frac{1}{\sum_{i=1}^G o_{\max}} 
\sum_{t=1}^{o_{\max}} \sum_{i=1}^G 
\frac{1}{\pi^{\text{emp}}_{\theta_{\text{old}}, t}} 
\left( \pi_{\theta}(o_{i,t} \mid q, o_{i,<t}) - \pi^{\text{emp}}_{\theta, t} \right) \hat{A}_i \right].
\label{eq:grpo_final2}
\end{aligned}
\end{equation}

From Eq.~\ref{eq:grpo_final2}, we observe that the model's update at each token position \( t \) is driven by the difference between the current policy's prediction and the empirical distribution, scaled by the trajectory's advantage \( \hat{A}_i \). This leads to the following interpretations:

\begin{itemize}
\item \textbf{Single-Trajectory View:} For a fixed trajectory $i$, the advantage $\hat{A}_i$ remains constant across all positions $t$. Thus, the model is encouraged to adjust the prediction $\pi_\theta(o_{i,t})$ at each token position proportionally to how much it deviates from the average distribution. Tokens with greater deviation from the average receive stronger updates, enabling the model to focus optimization on informative, distinctive tokens within the trajectory.

\item \textbf{Multi-Sample View:} When updates are aggregated from multiple trajectories, the signs of \( \hat{A}_i \) vary. This introduces a cancellation effect: tokens with larger deviations from the empirical distribution may contribute opposing gradients due to differing signs of \( \hat{A}_i \). Consequently, even if a position consistently demonstrates informative token choices, the gradient signals can be diminished or nullified due to interference across samples. This reduces the model's ability to reliably identify and leverage truly informative positions.
\end{itemize}

To mitigate this issue, we propose a \textbf{Token-Level Importance Weighting (TW)} strategy that selectively emphasizes positions exhibiting higher variability across sampled responses. The core insight is that token positions with greater distributional divergence indicate a mismatch between the current and optimal policies, suggesting these positions play a more influential role in model behavior. Our approach amplifies the learning signals associated with these informative tokens to counteract potential gradient cancellation effects, which enables more stable and effective policy optimization by preserving the influence of critical token positions.

\subsubsection{Token Importance via Information Content.} To more precisely quantify the informativeness of each token position, we introduce an information-theoretic measure: the \textbf{token-level information content}. Specifically, we propose to use the average Kullback-Leibler (KL) divergence across trajectories at each position $t$ to measure the degree of variation between individual token distributions and the expected distribution at that position. This provides a principled way to assess how “surprising” or “diverse” the predictions are at each step, which correlates with the position's importance to learning. Formally, we compute a token-level divergence score $D_t$ as:
\begin{equation}
D_t = \sum_{i=1}^{G} D_{\text{KL}}\left( p(o_{i,t}) \big\Vert \mathbb{E}[o_t] \right),
\end{equation}
where $G$ denotes the number of sampled outputs, $p(o_{i,t})$ is the probability distribution over tokens at position $t$ for trajectory $i$, and $\mathbb{E}[o_t]$ is the expected distribution at position $t$, estimated by averaging the predicted distributions across all trajectories. To accommodate variable-length sequences, any missing tokens are filled using a uniform distribution $\mathcal{U}(V)$ over the vocabulary $V$. This ensures all positions are comparably represented, and prevents bias toward longer sequences.

To normalize across positions and maintain stable optimization, we apply min-max normalization:
\begin{equation}
w_t = (1 + \alpha) \cdot \frac{D_t - D_{\min}}{D_{\max} - D_{\min}},
\end{equation}
where $\alpha$ is a hyperparameter controlling the baseline importance of low-divergence positions. The additive constant $(1 + \alpha)$ ensures that no token position receives a zero weight, thus preventing complete gradient suppression at any location.

\subsubsection{Final Objective with Token-Level Importance Weighting}
We integrate these token-level weights into Eq.~\ref{eq:dapo_loss_ap} to form the Token-Level Importance Weighting GRPO (TW-GRPO) objective:
\begin{equation}
\begin{aligned}
\mathcal{J}_{\text{TW-GRPO}}(\theta) 
&=  \mathbb{E}_{(q,a)\sim \mathcal{D}, \{o_i\}_{i=1}^G \sim \pi_{\theta_\text{old}}(\cdot \mid q)} \\
&\Bigg[ \frac{1}{\sum_{i=1}^G |o_i|} 
\sum_{i=1}^G \sum_{t=1}^{|o_i|} 
\min \Big( w_t \cdot r_{i,t}(\theta) \hat{A}_{i,t}, \ \text{clip} \big( r_{i,t}(\theta), 1 - \varepsilon, 1 + \varepsilon \big) \hat{A}_{i,t} \Big) \Bigg].
\end{aligned}
\label{eq:tw_grpo_loss_ap}
\end{equation}

\section{Detailed Experimental Setup}

\label{sec:detailed_setup}


\label{evaluation_setting}
Evaluation is conducted on six video benchmarks covering general understanding and reasoning: MVBench~\cite{li2024mvbench}, TempCompass~\cite{liu2024tempcompass}, VideoMME~\cite{fu2024video}, MMVU~\cite{zhao2025mmvu}, NExT-GQA~\cite{0Can}, and CLEVRER~\cite{yi2019clevrer}. MVBench, TempCompass, and VideoMME emphasize general video understanding, combining visual perception and temporal comprehension without explicit reasoning focus. CLEVRER, NExT-GQA, and MMVU assess complex spatiotemporal and multimodal reasoning over dynamic videos. Together, these benchmarks comprehensively evaluate both general video understanding and fine-grained multimodal reasoning capabilities, ensuring a well-rounded assessment of the model’s performance.

For all evaluations, we followed the decoding configuration used in the official Qwen2.5-VL demo, with top\_p = 0.001 and temperature = 0.01. We adopted the same experimental settings as in Video- R1~\cite{videor1}, including the sampling temperature and top\_p, and set the batch size to 16. For the NExT-GQA, MMVU, MVBench, TempCompass, and VideoMME datasets, we used the prompt words from Video R1. For the CLEVRER dataset, we adopt the simple prompting strategy designed for our indefinite-choice setting:  
\textit{"Output the thinking process in <think></think> and the final answer (letters separated by commas, if multiple) in <answer></answer> tags."}  
In particular, we evaluate and train on its most challenging subset—counterfactual questions. For the other benchmarks, we follow the evaluation setup of Video-R1, conducting experiments on a partial subset of VideoMME and the multiple-choice question split of MMVU.

\section{Additional Experiment Results}

\subsection{Effect of Sampling Diversity on Approximation Validity}
\label{ass_exp}

\begin{table*}[t]
\centering
\renewcommand{\thetable}{A\arabic{table}}  
\setcounter{table}{0} 
\caption{\textbf{Sensitivity analysis on sampling number and temperature on CLEVRER and MMVU datasets.} Metrics include accuracy and soft accuracy under different settings.}
\label{tab:sensitivity}
\vspace{1mm}
\renewcommand{\arraystretch}{1.1}
\setlength{\tabcolsep}{1.5mm}
\begin{tabular}{l|c|cc|cc|c}
\toprule
\multirow{2}{*}{\textbf{Setting}} & \textbf{Single.} & \multicolumn{2}{c|}{\textbf{Multiple Choice}} & \multicolumn{2}{c|}{\textbf{All}} & \textbf{MMVU} \\
& Acc. (\%) & Acc. (\%) & Soft Acc. (\%) & Acc. (\%) & Soft Acc. (\%) & Acc. (\%) \\
\midrule
\multicolumn{7}{l}{\textit{\textbf{Sampling Number}}} \\
4   & 56.2 & 26.0 & 55.2 & 38.9 & 55.6 & \textbf{65.9} \\
\rowcolor{gray!5} 8   & 60.9 & \textbf{42.5} & \textbf{64.4} & \textbf{50.4} & 62.9 & 65.8 \\
12  & \textbf{65.1} & 37.2 & 62.8 & 49.1 & \textbf{63.8} & 63.2 \\
\midrule
\multicolumn{7}{l}{\textit{\textbf{Temperature}}} \\
0.5 & 63.4 & 33.5 & 61.0 & 46.3 & 62.1 & 65.3 \\
\rowcolor{gray!5} 1.0 & 60.9 & \textbf{42.5} & \textbf{64.4} & \textbf{50.4} & \textbf{62.9} & \textbf{65.8} \\
1.5 & 51.7 & 38.7 & 61.5 & 44.3 & 57.4 & 63.7 \\
\bottomrule
\end{tabular}
\vspace{-4mm}
\end{table*}

As shown in Table~\ref{tab:sensitivity}, we conduct a sensitivity analysis on two key generation hyperparameters: the number of sampled trajectories and the decoding temperature. These factors directly influence the validity of Assumption~\ref{assumption}, which requires (i) a sufficiently large but not excessively large sampling size, and (ii) the old policy to behave consistently across similar contexts.

We observe that when the number of sampled trajectories is too small (e.g., 4), model performance degrades notably. This is expected, as a small sample size violates the assumption of empirical distribution reliability: it introduces high variance into the denominator approximation in importance weighting. Conversely, when the number of samples is too large (e.g., 12), performance again drops, likely due to increased inter-sample variability—larger sample sizes lead to more diverse histories $o_{j,<t}$, thereby increasing the heterogeneity in $\pi_{\theta_{\text{old}}}(\cdot \mid q, o_{j,<t})$, and breaking the assumption of policy stability across contexts.

Temperature exhibits a similar trade-off. When the temperature is too low (e.g., 0.5), model outputs become overly deterministic, reducing the diversity needed for advantage computation to distinguish informative tokens via deviation. On the other hand, very high temperatures (e.g., 1.5) induce excessive randomness, which again leads to large variability across trajectories and violates the assumption of stability in the policy's behavior.

In summary, both the sample size and the generation temperature must be carefully tuned to maintain a balance: enough diversity to support optimization via sample-level deviation, but not so much that it invalidates the Assumption~\ref{assumption}. This results empirically support the necessity of Assumption~\ref{assumption} for TW-GRPO to be effective.

\subsection{Sensitivity Analysis on Token-Level Importance Weighting Mechanism}

\label{sec:sensitivity}

We conduct sensitivity studies to better understand how the design of the TW-GRPO mechanism affects model performance. In particular, we focus on two key components: (i) the weight coefficient $\alpha$, and (ii) the positional scope over which token-level importance weighting is applied.

\subsubsection{Effect of Weight Coefficient $\alpha$}

We analyze the impact of the TW-GRPO weighting hyperparameter $\alpha$ on model performance. As shown in Figure~\ref{fig:alpha_analysis}, the results across different datasets exhibit similar trends: performance improves as $\alpha$ increases up to a point and then begins to degrade. Most datasets achieve optimal results around $\alpha = 0.7$, with exceptions such as VideoMME, where $\alpha = 0.5$ yields the best outcome.

These results suggest that smaller $\alpha$ values under-emphasize the contribution of important tokens during optimization, limiting the potential of TW-GRPO. On the other hand, excessively large $\alpha$ values lead to over-concentration on high-deviation tokens, potentially ignoring broader contextual information—analogous to the “blind men and the elephant” problem. Thus, careful tuning of $\alpha$ is critical; we use $\alpha = 0.7$ as the default setting in all main experiments.

\renewcommand{\thefigure}{A\arabic{figure}}
\setcounter{figure}{0}

\begin{figure*}
    \centering
    \includegraphics[width=1\textwidth]{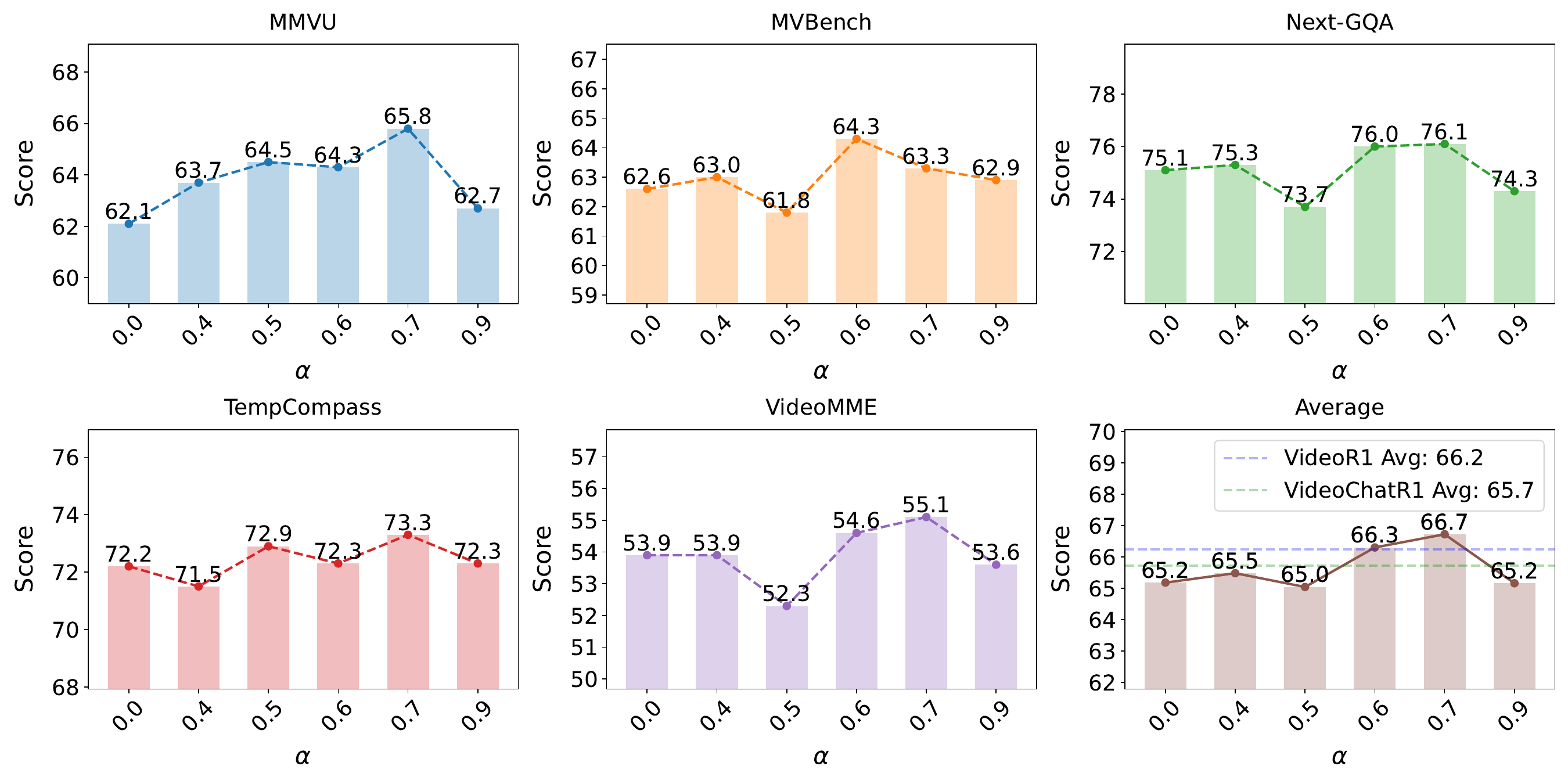} 
    \caption{Analysis of the influence of the TW-GRPO weighting coefficient $\alpha$ on model performance.}
    \label{fig:alpha_analysis}
    \vspace{-6mm}
\end{figure*}

\subsubsection{Effect of Weighting Token Positions}

\begin{figure*}
    \centering
    \includegraphics[width=1\textwidth]{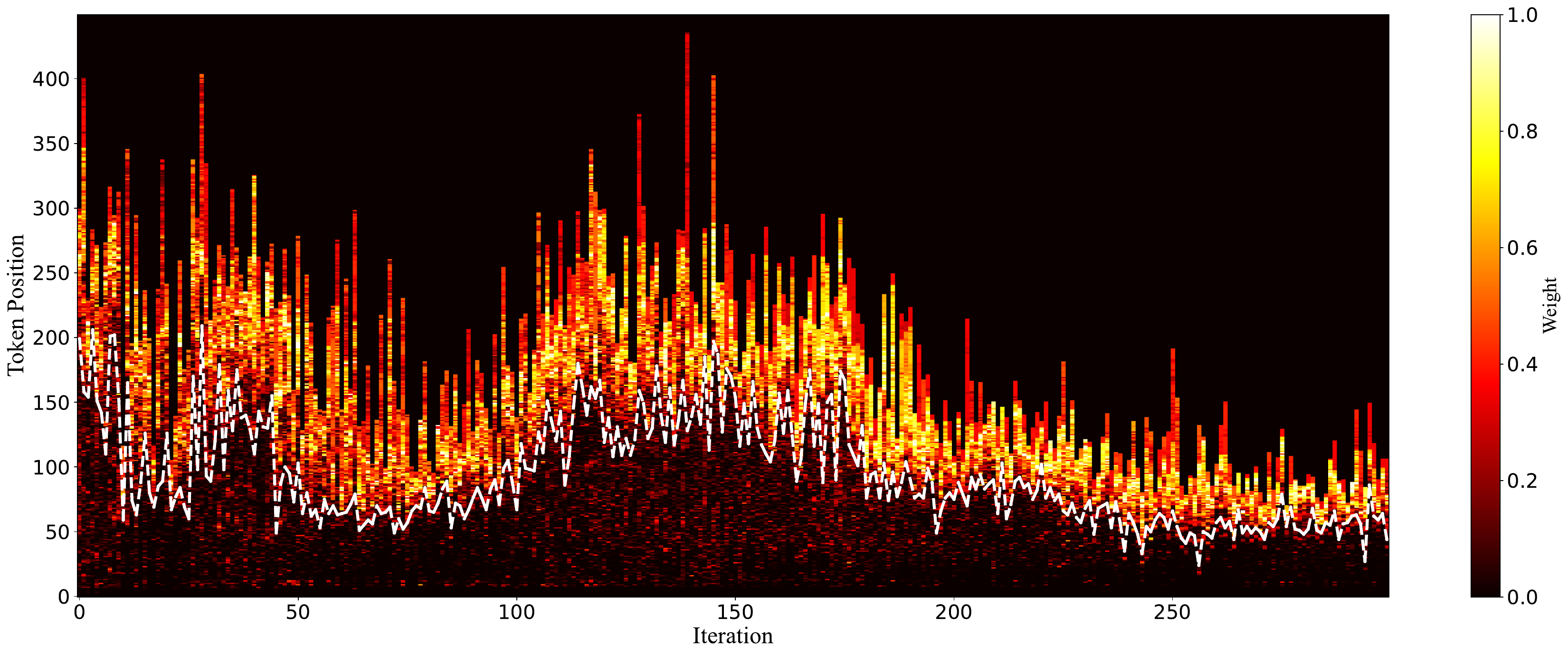} 
    \caption{Visualization of token-level importance weights across all positions on the first 300 training samples. The area above the white dashed line corresponds to padding tokens.}
    \label{fig:all_postions}
    \vspace{-6mm}
\end{figure*}

\renewcommand{\thetable}{A\arabic{table}}
\setcounter{table}{1}  
\begin{table*}[t]
\caption{Performance of TW-GRPO under different token-weighting position strategies across video reasoning and general benchmarks.}
\label{tab:weighted_position_main}
\resizebox{\linewidth}{!}{%
\setlength{\tabcolsep}{0.8mm}
\renewcommand\arraystretch{1.3}
\begin{tabular}{@{}lcccc|ccc@{}}
\toprule
\multirow{2}{*}{\textbf{Weighting Position}} & \multicolumn{3}{c}{\textbf{Video Reasoning Benchmark}} & \multicolumn{3}{c}{\textbf{Video General Benchmark}} \\ 
\cmidrule(l){2-7} 
& CLEVRER$\rm_{cf}$ & NExT-GQA & MMVU$\rm _{mc}$ & MVBench & TempCompass & VideoMME$\rm _{(wo \ sub)}$ \\ 
\midrule
None (No Weighting)         & 48.6 & 75.1 & 62.1 & 62.6 & 72.2 & 53.9 \\
Padding Only                & 47.0 & 74.7 & 64.3 & 62.7 & 72.6 & 53.7 \\
Content Only                & 46.0 & \textbf{76.1} & \textbf{66.2} & \textbf{63.4} & 72.7 & \textbf{55.7} \\
\rowcolor{gray!5} All Positions              & \textbf{50.4} & \textbf{76.1} & {65.8} & {63.3} & \textbf{73.3} & {55.1} \\
\bottomrule
\end{tabular}
}
\vspace{-4mm}
\end{table*}

In practice, since the model samples multiple completions per prompt and each sampled response may vary in length, we apply padding to align all sequences to a uniform maximum length. This enables consistent token-wise operations, such as weighting, across different completions. In this section, we evaluate the impact of token-level importance weighting positions:
(i) None, which applies no weighting to any token positions; (ii) Padding Only, which applies weighting exclusively to the padded token postions introduced to match sequence lengths; (iii) Content Only, which applies weighting only to positions that contain valid content across all completions (i.e., tokens that are present in every sampled sequence before padding); (iv) All Positions, which applies weighting uniformly to all token positions, including both content and padding. 

Table~\ref{tab:weighted_position_main} shows that the All Positions configuration consistently achieves the best performance across all datasets. It provides substantial gains on CLEVRER, MMVU, and VideoMME, indicating that the token-level importance signals captured by TW-GRPO are beneficial in padding regions and throughout the full token sequence. This suggests that incorporating signals from all token positions allows for more robust optimization, especially in settings with variable-length outputs.

The Content Only strategy performs competitively on several benchmarks, such as NExT-GQA, MMVU, and VideoMME, and even outperforms other strategies in some cases. This suggests that focusing solely on content postions helps capture critical reasoning information, even though the importance weights assigned to content tokens are significantly lower than those involving padding tokens, as shown in Figure~\ref{fig:all_postions} . However, its performance on CLEVRER is inferior to that of All Postions, suggesting that excluding padding may overlook useful auxiliary signals. In cases with variable output lengths, padding positions may carry important alignment or reasoning-related information, especially when short responses are padded to match longer ones.

As shown in Figure~\ref{fig:all_postions} , although tokens in padding regions often receive relatively high importance weights, the Padding Only strategy yields only slight improvements over None on a few benchmarks, such as MMVU, and remains consistently less effective than the All Positions configuration. This suggests that restricting the weighting to padding positions fails to capture critical information embedded in the content regions, which is more effective in comparison, as shown in Table~\ref{tab:weighted_position_main}, resulting in limiting the overall effectiveness of token-level optimisation.

Overall, the consistent improvements obtained with All Postions confirm the importance of a global token-weighting mechanism for maximizing TW-GRPO performance. In terms of token-level distinction, TW-GRPO assigns near-zero importance to initial tokens such as generic openings like “The video shows...”, which appear consistently across samples and contribute little to task-specific reasoning. Their low variance and minimal distributional divergence naturally lead to low importance weights. In contrast, tokens in the middle of the sequence exhibit significantly higher and more variable weights. These positions often correspond to reasoning-critical content, including causal links, temporal dynamics, or attribute comparisons, which vary substantially across completions. This indicates that TW-GRPO can effectively identify semantically meaningful token-level differences and direct optimization toward the most informative regions. By enabling finer granularity and position-aware gradient updates, the model benefits from all available token-level signals, leading to overall performance gains.

\subsubsection{Analysis of $D_t$ During Training}

\begin{figure*}
    \centering
    \includegraphics[width=1\textwidth]{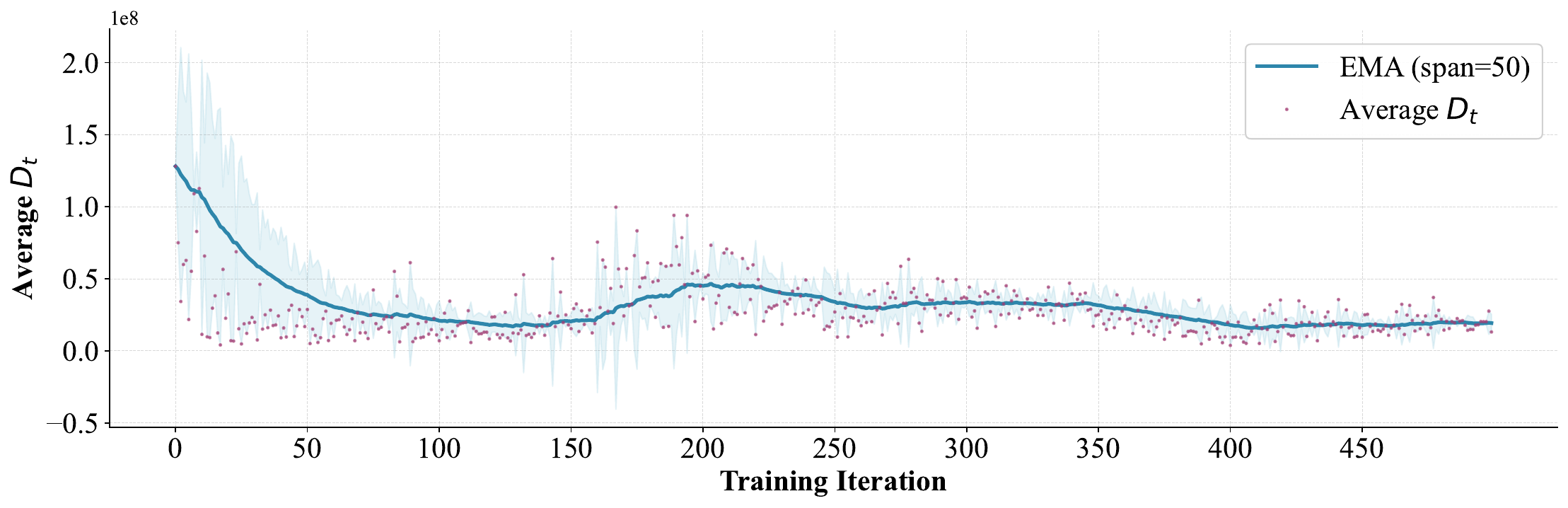} 
    \caption{Visualization of $D_t$ across Content Only on the First 500 Training Samples.}
    \label{fig:competion_only}
\vspace{-6mm}
\end{figure*}

To further understand how TW-GRPO leverages token-level information throughout training, we visualize the evolution of the average divergence score $D_t$ across iterations, as shown in Fig.~\ref{fig:competion_only}. As training progresses, $D_t$ gradually flattens, indicating a decline in variance across token distributions. This trend reflects a core intuition: when a model becomes more confident, it generates increasingly consistent token-level predictions across samples. Such convergence in output distributions, as quantified by the declining $D_t$, has been shown to align with improved reasoning performance in prior work~\cite{agarwal2025unreasonable, zhang2025right, zhao2025learning}, where deterministic or low-entropy behavior is positively correlated with predictive accuracy and generalization. To encourage this desirable behavior during training, TW-GRPO explicitly incorporates token-level uncertainty into the optimization process. Our method uses KL divergence across samples to identify uncertain or unstable token positions. High divergence indicates locations where the model remains uncertain, and TW-GRPO amplifies learning signals at these positions. As training continues and these tokens become more stable, their weights naturally decrease, resulting in a smooth convergence of $D_t$. This dynamic behavior facilitates broad exploration during early training and enables more focused convergence in later stages, thereby validating the effectiveness of our token-level importance weighting strategy in enhancing reasoning optimization.

\section{Additional Visualization Results}
\label{visualization}

\subsection{Analysis of Reasoning Path}
\begin{figure}[t]
  \centering
  \includegraphics[width=1\textwidth]{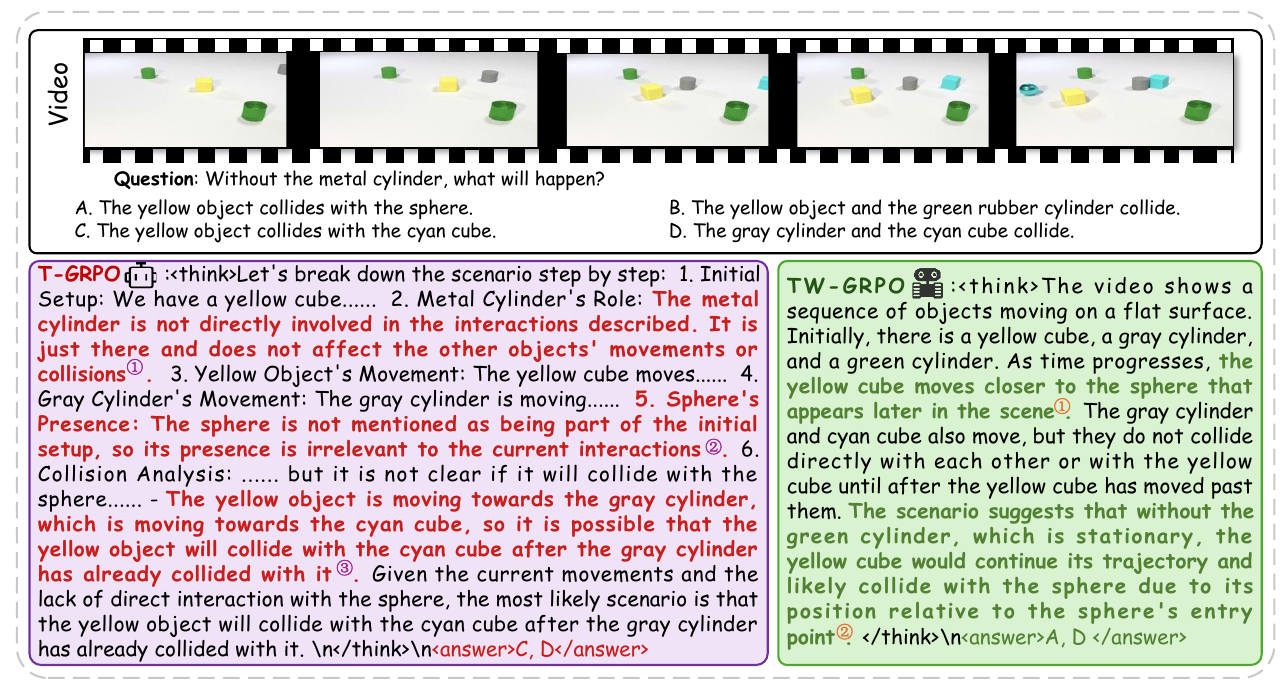} 
  \caption{Comparison of reasoning paths from T-GRPO and TW-GRPO on CLEVRER samples. TW-GRPO accurately reasons about dynamically introduced objects and counterfactual outcomes (e.g., object removal), enabling stronger causal reasoning.}
  \label{fig:comparison example1}
  \vspace{-4mm}
\end{figure}

\begin{figure}[t]
  \centering
  \includegraphics[width=1\textwidth]{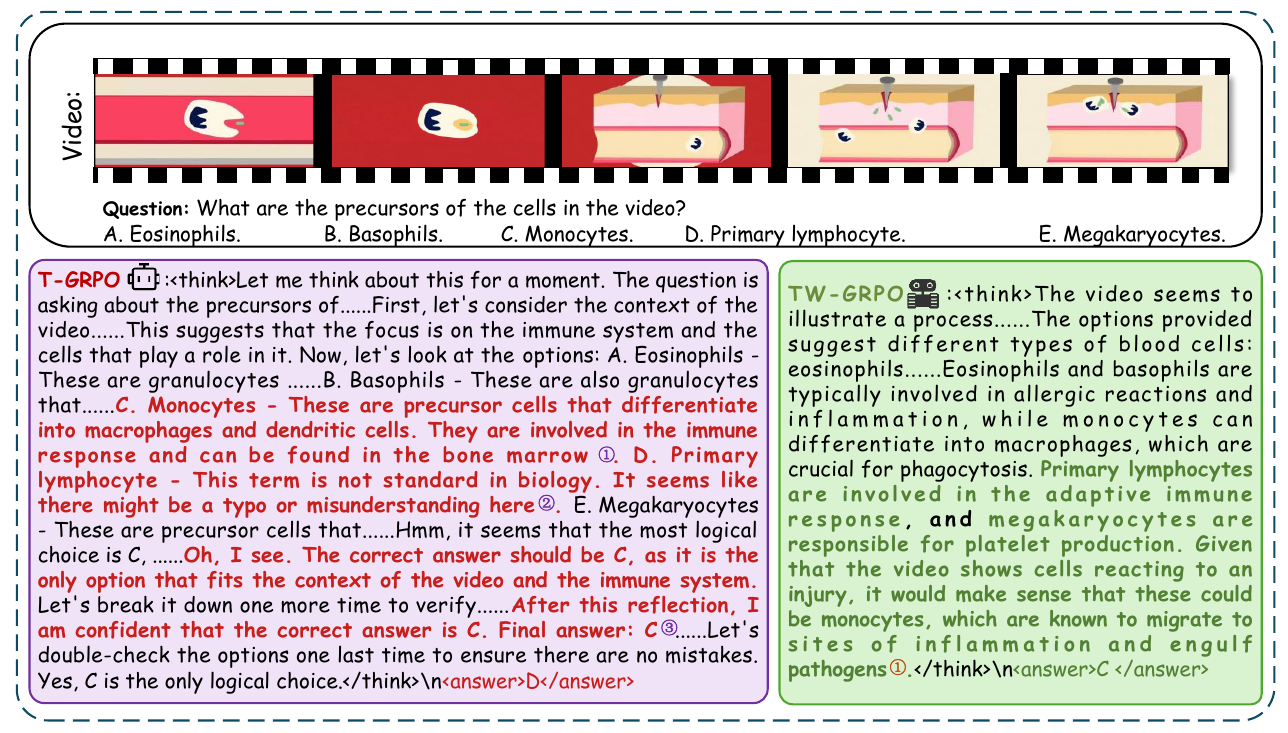} 
  \caption{Comparison of reasoning paths from T-GRPO and TW-GRPO on MMVU samples. TW-GRPO achieves more accurate conclusions by leveraging stronger causal reasoning and better alignment with visual and medical knowledge.}
  \label{fig:comparison example2}
  \vspace{-6mm}
\end{figure}

We compare models trained with T-GRPO and TW-GRPO on representative samples from the CLEVRER and MMVU datasets. In Figure~\ref{fig:comparison example1}, the model trained with T-GRPO focuses on local object motion but misses broader contextual factors such as the blocking role of the metal cylinder \circled[purple]{1} and the dynamically introduced sphere \circled[purple]{2}. It does not simulate hypothetical changes like object removal \circled[purple]{3}, which limits the scope of its prediction. The TW-GRPO-trained model, in contrast, integrates the late-appearing sphere into its reasoning \circled[orange]{1} and simulates the removal of the green cylinder to infer its effect on future trajectories \circled[orange]{2}, demonstrating stronger temporal and causal understanding. A similar pattern appears in Figure~\ref{fig:comparison example2}, the T-GRPO trained model initially reasons about monocytes correctly \circled[purple]{1}, but exhibits inconsistency when evaluating distractor options \circled[purple]{2}, \circled[purple]{3}. TW-GRPO maintains a coherent reasoning chain by aligning immune activity with monocyte function and contextual visual cues \circled[orange]{1}. These examples highlight how TW-GRPO enhances the accuracy of reasoning by integrating dynamic, contextual, and causal relationship, leading to more precise and reliable conclusions.

\section{Limitations}
Although our method demonstrates strong performance across multiple tasks, it still has several limitations. First, due to considerations of training efficiency and computational resources, we follow the same training and evaluation settings as Video-R1 by uniformly sampling each video into 16 frames. However, the frame sampling strategy used in the VideoChat-R1 paper differs from ours. To address this, we utilize the publicly released weights of VideoChat-R1 and report the test results under their original settings. For transparency and reproducibility, we provide the full benchmark results and corresponding evaluation logs for both our method and VideoChat-R1 in the anonymous repository. Second, the proposed Question-Answer Inversion method currently relies on direct string matching, which requires inputs to conform to a specific format. This constraint may limit its generalizability to more diverse or unstructured question-answer pairs. Finally, while our optimization approach achieves strong empirical results, it relies on Assumption~\ref{assumption} being satisfied. As a result, it imposes certain requirements on the sampling configuration of the model, particularly the number of sampled responses and the temperature setting during inference.